\documentclass[sigconf]{acmart}
\usepackage{xspace}

\usepackage{multirow}
\usepackage{colortbl}
\usepackage{booktabs}
\IfFileExists{libertine.sty}{}{\microtypesetup{expansion=false}}
\graphicspath{{./}{samples/}}
\definecolor{tabfirst}{rgb}{1, 0.6, 0.6} 
\definecolor{tabsecond}{rgb}{1, 0.8, 0.5} 
\definecolor{tabthird}{rgb}{1, 1, 0.6} 
\definecolor{lightred}{rgb}{1, 0.6, 0.6}
\definecolor{lightorange}{rgb}{1, 0.8, 0.5}
\definecolor{lightyellow}{rgb}{1, 1, 0.6}

\acmConference[ACM MM '26]{ACM Multimedia 2026}{November 10--14, 2026}{Rio de Janeiro, Brazil }

\setcopyright{none}
\renewcommand\footnotetextcopyrightpermission[1]{}

\begin{document}
\raggedbottom
\setlength{\emergencystretch}{2em}

\title{SalientGS: Unified SfM-to-3DGS with Importance-Guided MCMC Gaussian Allocation}

\author{Tianyu Xiong}
\affiliation{%
  \institution{School of Computer Science, Northwestern Polytechnical University}
  \city{Xi'an}
  \country{China}
}
\author{Rui Li}
\affiliation{%
  \institution{CEMSE, King Abdullah University of Science and Technology}
  \city{Thuwal}
  \country{Saudi Arabia}
}

\author{Suning Ge}
\affiliation{%
  \institution{School of Computer Science, Northwestern Polytechnical University}
  \city{Xi'an}
  \country{China}
}

\author{Jiaqi Yang}
\affiliation{%
  \institution{School of Computer Science, Northwestern Polytechnical University}
  \city{Xi'an}
  \country{China}
}

\renewcommand{\shortauthors}{T. Xiong et al.}

\begin{abstract}
Reconstructing 3D scenes from unordered images remains bottlenecked by expensive Structure-from-Motion (SfM) preprocessing and frozen pose interfaces. We present SalientGS, a unified SfM-to-3D Gaussian Splatting (3DGS) pipeline. Its central contribution is importance-guided Markov Chain Monte Carlo (MCMC) Gaussian allocation, which aggregates multi-view residuals into per-Gaussian underfit and redundancy signals. These signals define a smooth importance-weighted sampling distribution that biases both birth and relocation toward underfit regions. This reallocates capacity from well-fit areas without altering the underlying stochastic gradient Langevin dynamics (SGLD). In a released-code verification over 13 scenes and three benchmarks, SalientGS achieves the best cross-benchmark macro-average PSNR, SSIM, and LPIPS (27.65~dB / 0.876 / 0.147) among the compared methods, while also providing the fastest end-to-end runtime (10.62 minutes) with 1.5M Gaussians. Code, per-scene measurements, and evaluation scripts are available at \url{https://github.com/Six-Bit-TX/SalientGS}.
\end{abstract}



\maketitle

\begin{figure}[!t]
    \centering
    \includegraphics[width=\columnwidth]{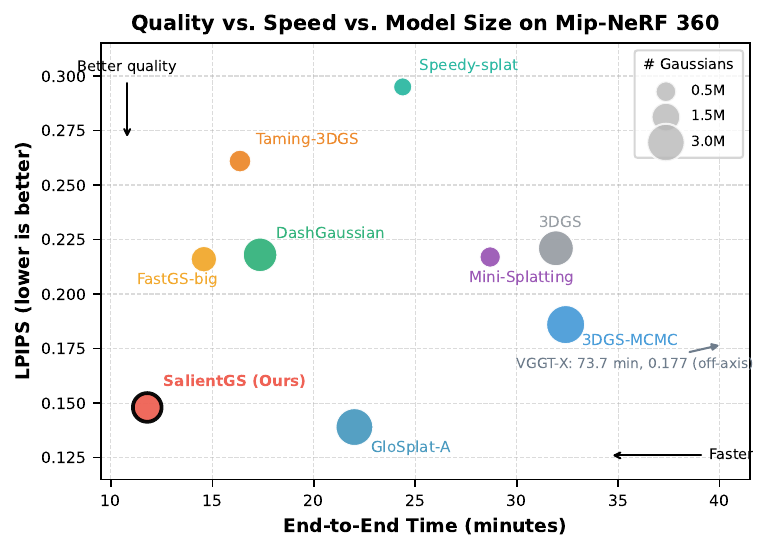}
    \caption{\textbf{Quality vs.\ Speed vs.\ Model Size on Mip-NeRF 360.} Bubble size indicates Gaussian count. SalientGS combines strong perceptual quality with the fastest end-to-end runtime in this comparison, using 1.5M Gaussians and no standalone COLMAP preprocessing.}
    \label{fig:teaser}
\end{figure}

\begin{figure*}[t]
    \centering
    \includegraphics[width=\textwidth]{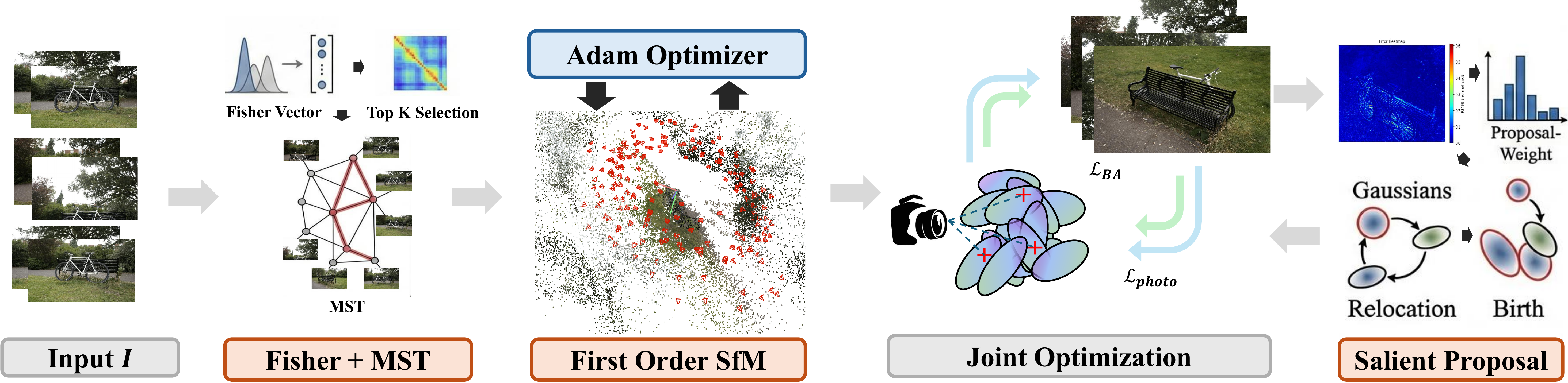}
    \caption{Starting from an unordered image set, we first retrieve candidate pairs with Fisher Vector descriptors and enforce global connectivity with a maximum spanning tree to build a sparse but reliable matching graph. We then run a first-order SfM stage to estimate initial camera poses and triangulate a coarse 3D structure, followed by joint optimization of the 3D Gaussian representation and camera parameters under photometric and reprojection-based BA losses. This unified pipeline concentrates model capacity on salient scene regions while preserving geometric anchoring throughout optimization. Red marks indicate the main novel components of our method.}
    \label{fig:pipeline}
\end{figure*}

\section{Introduction}
\label{sec:introduction}

Reconstructing high-fidelity 3D scenes from unordered image collections is a fundamental problem in computer vision and graphics, with applications spanning virtual reality, robotics, and cultural heritage preservation. The recent advent of 3D Gaussian Splatting (3DGS)~\cite{kerbl3Dgaussians} has transformed this landscape, enabling real-time novel view synthesis with quality rivaling Neural Radiance Fields (NeRF)~\cite{mildenhall2021nerf} while dramatically reducing rendering time. However, 3DGS inherits a critical dependency from the NeRF paradigm: it requires accurate camera poses and sparse point clouds from Structure-from-Motion (SfM) preprocessing, most commonly performed with COLMAP~\cite{colmap}.

In widely used COLMAP-based two-stage pipelines, this preprocessing step can become a practical bottleneck for end-to-end reconstruction. Exhaustive image matching and second-order bundle adjustment can be expensive, especially for large image collections, and their cost can rival or exceed 3DGS training time itself (Table~\ref{tab:main-result}). More importantly, the interface is typically frozen: SfM pose errors propagate downstream without photometric correction, and the separation between stages limits joint optimization of geometry and appearance.

Recent efforts to accelerate 3DGS training~\cite{mallick2024taming,chen2025dashgaussian,fang2024mini,ren2025fastgs} have achieved impressive speedups through improved densification strategies, progressive training, and efficient Gaussian management. However, these methods still rely on standalone COLMAP preprocessing, meaning their reported times understate the true end-to-end cost. Meanwhile, COLMAP-free approaches~\cite{fu2024colmap,ji2025sfm,huang20253r} have emerged, but many assume sequential input, use feed-forward networks that can be sensitive to domain shift, or omit geometric constraints during training, leading to pose drift in challenging scenarios.

We present \textbf{SalientGS}, a unified SfM-to-3DGS pipeline that combines fast reconstruction with competitive rendering quality under a fixed 1.5M-Gaussian budget. Our primary algorithmic contribution is \emph{importance-guided MCMC Gaussian allocation}. We compute multi-view underfit (importance) and well-fit (redundancy) signals, then convert them into an importance-weighted sampling distribution. This distribution biases \textbf{both} birth and relocation within the MCMC framework, reallocating a fixed Gaussian budget from redundant regions to persistent errors. Crucially, this importance guidance operates as a heuristic allocation strategy layered on top of the SGLD-based population dynamics of 3DGS-MCMC~\cite{mcmc3dgs}; it does not alter the underlying Langevin updates and therefore makes no additional convergence claims beyond those of the base framework. Our key insight is that \emph{importance-guided capacity reallocation} makes a fast but coarse SfM initialization viable when it is coupled with joint refinement of pose and appearance and with geometric anchoring.

Unlike prior joint optimization methods that rely primarily on photometric gradients, SalientGS maintains explicit geometric constraints through a reprojection-based bundle adjustment (BA) loss on triangulated feature tracks, enabling accurate pose refinement while preventing degenerate solutions.
Our contributions:
{\setlength{\topsep}{2pt}
\setlength{\itemsep}{2pt}
\setlength{\parsep}{0pt}
\setlength{\partopsep}{0pt}
\begin{itemize}
    \item \textbf{Importance-guided MCMC Gaussian allocation:}\newline A heuristic allocation layer atop 3DGS-MCMC~\cite{mcmc3dgs} that biases both birth and relocation toward underfit regions via multi-view error attribution, yielding +0.10~dB PSNR and a 0.001 LPIPS reduction over vanilla MCMC at 1.5M Gaussians (Table~\ref{tab:ablation}).
    \item \textbf{Unified SfM-to-3DGS pipeline:} An end-to-end architecture coupling fast SfM initialization with joint refinement of pose and appearance under photometric and reprojection losses.
    \item \textbf{Efficient matching and first-order SfM:} Fisher Vector (FV) retrieval with Maximum Spanning Tree (MST) connectivity and first-order epipolar adjustment, achieving near-linear scaling and up to 23$\times$ SfM speedup.
    \item \textbf{Released-code verification:} A reproducible 13-scene evaluation with exact 30K-step schedules and per-scene records; SalientGS gives the best three-benchmark macro-average PSNR/SSIM/LPIPS and runtime among all methods in Table~\ref{tab:main-result}.
\end{itemize}
}

\section{Related Work}
\label{sec:related_work}

\subsection{Novel View Synthesis and 3D Gaussian Splatting}

NeRF~\cite{mildenhall2021nerf} represents scenes as continuous volumetric functions, with improvements in anti-aliasing~\cite{barron2021mip,barron2022mip} and training speed~\cite{muller2022instant}. 3DGS~\cite{kerbl3Dgaussians} enables real-time rendering via anisotropic Gaussians but is sensitive to initialization quality. Recent acceleration efforts include importance-based budgeting~\cite{mallick2024taming}, progressive reconstruction~\cite{chen2025dashgaussian}, pruning~\cite{fang2024mini,hanson2025speedy}, and MCMC-based management~\cite{mcmc3dgs} with SGLD-driven birth-death processes. FastGS~\cite{ren2025fastgs} combines many of these innovations. We adopt the MCMC framework and layer \emph{importance-guided} birth and relocation on top, biasing allocation via multi-view error attribution without modifying the underlying SGLD dynamics.

\subsection{Structure from Motion}

Incremental SfM pipelines such as COLMAP~\cite{colmap} are widely used and robust, but they can accumulate drift and often rely on computationally heavy matching and second-order BA. Global SfM methods~\cite{moulon2017openmvg,sweeney2015theia} solve all poses simultaneously via rotation~\cite{hartley2013rotation,wilson2020distribution} and translation averaging~\cite{govindu2001combining,martinec2007robust}; GLOMAP~\cite{pan2024glomap} achieves COLMAP-level accuracy with significant speedups. FastMap~\cite{fastmap} replaces second-order BA with first-order structureless epipolar adjustment, achieving BA-quality refinement with cost independent of point count. Learning-based approaches include differentiable SfM~\cite{wang2024vggsfm}, flow-based optimization~\cite{smith2024flowmap}, and the DUSt3R/MASt3R/VGGT paradigm~\cite{wang2024dust3r,leroy2024grounding,wang2025vggt}. We integrate first-order SfM with joint 3DGS training, retaining an optimization-based geometric backbone while avoiding feed-forward domain sensitivity.

\subsection{Image Matching and Retrieval}

Exhaustive pairwise matching scales as $O(N^2)$. Retrieval-based pair selection addresses this via image-level descriptors such as Fisher Vectors~\cite{perronnin2010fisher}, Bag-of-Words~\cite{sivic2003video,jegou2010aggregating}, and learned methods like NetVLAD~\cite{arandjelovic2016netvlad} and MegaLoc~\cite{megaloc}. We adopt Fisher Vector retrieval with MST-based connectivity guarantees to achieve near-linear matching complexity without requiring pretrained networks.

\subsection{Joint Pose and Appearance Optimization}

Traditional pipelines treat SfM and novel view synthesis as independent modules with frozen interfaces. NeRF--~\cite{wang2021nerfmm} and BARF~\cite{lin2021barf} optimize poses via photometric gradients; SPARF~\cite{truong2023sparf} adds multi-view correspondences but lacks explicit geometric constraints. COLMAP-free 3DGS methods have also emerged: CF-3DGS~\cite{fu2024colmap} and HT-3DGS~\cite{ji2025sfm} assume sequential input, 3RGS~\cite{huang20253r} relies on photometric-only refinement, and GloSplat~\cite{glosplat} preserves SfM feature tracks for geometric anchoring. Unlike these methods, we maintain geometric constraints through a reprojection-based BA loss on triangulated tracks within a unified end-to-end pipeline.

\section{Problem Description and Methodology}
\label{sec:methodology}

Given $N$ unordered images depicting a scene, we simultaneously reconstruct a 3DGS representation and recover camera poses through three stages (Figure~\ref{fig:pipeline}): global matching, first-order SfM, and joint 3DGS training with importance-guided MCMC allocation.

\subsection{Global Correspondence with Fisher Vector \& MST}
\label{sec:global_matching}

\begin{figure}[t]
    \centering
    \includegraphics[width=\columnwidth]{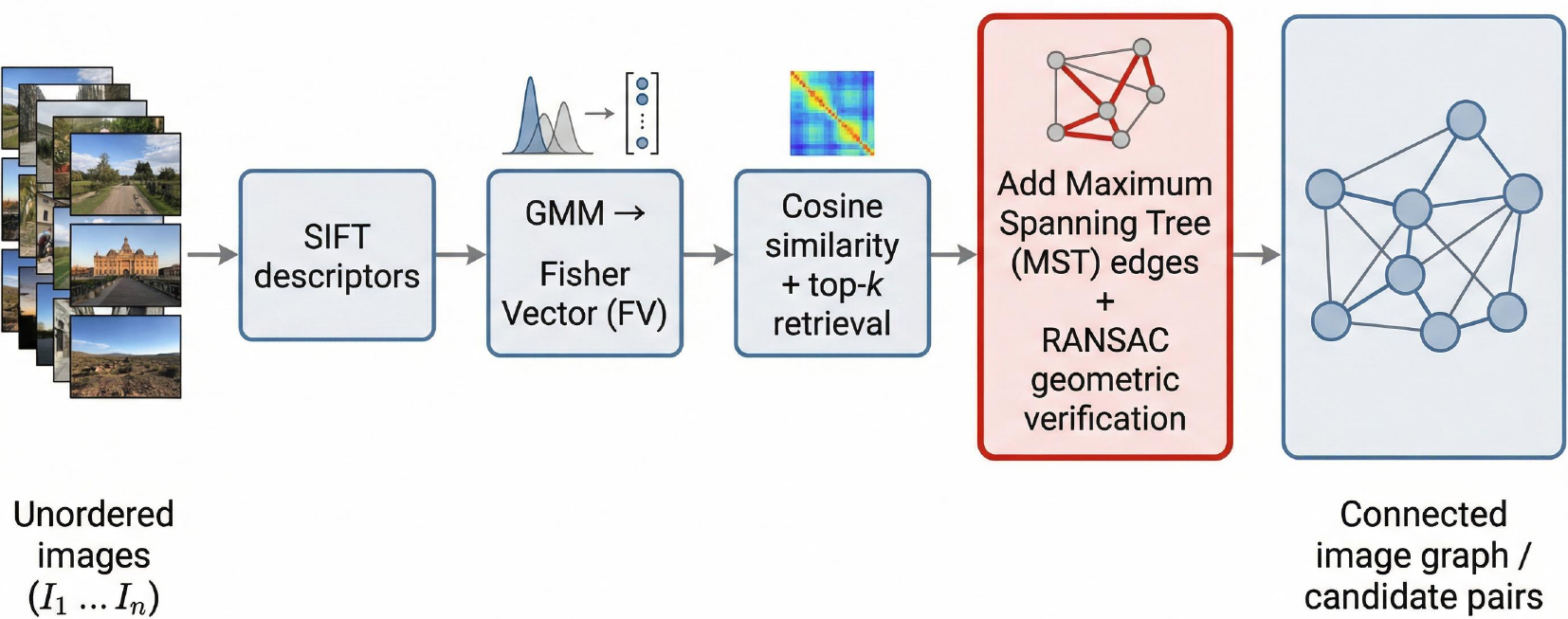}
    \caption{Given an unordered image collection, we extract Scale-Invariant Feature Transform (SIFT) features, encode them into Fisher Vectors, retrieve top-$k$ candidate neighbors, add MST edges, and verify pairs with RANSAC. This produces a sparse but well-connected matching graph that balances reconstruction quality and efficiency.}
    \label{fig:fisher_mst}
\end{figure}

Traditional exhaustive image matching scales as $O(N^2)$, becoming prohibitive for large image collections. We address this through a retrieval-based approach that identifies visually similar image pairs using Fisher Vector (FV) global descriptors, followed by Maximum Spanning Tree (MST) connectivity guarantees. Figure~\ref{fig:fisher_mst} illustrates our retrieval and pair selection pipeline.

\paragraph{Fisher Vector Encoding}
For each image $I_i$, we extract Scale-Invariant Feature Transform (SIFT) descriptors and encode them into a Fisher Vector $\mathbf{f}_i$~\cite{perronnin2010fisher} by computing gradients of the descriptor log-likelihood with respect to an $M$-component Gaussian Mixture Model (GMM) trained on the image collection. The resulting vectors are L2- and signed-square-root normalized for retrieval.

\paragraph{Pair Selection with MST Connectivity}
We retrieve the top-$k$ most similar images per query via FAISS~\cite{faiss} approximate nearest-neighbor (ANN) search in $O(N \log N)$ time, producing a sparse $k$NN graph. To guarantee connectivity for global SfM, we add Maximum Spanning Tree edges from this sparse graph, then apply RANSAC-based geometric verification to filter spurious matches. The final pair set is $\mathcal{P} = \mathcal{P}_{\text{top-}k} \cup \mathcal{P}_{\text{MST}} \cup \mathcal{P}_{\text{extra}}$.

\subsection{First-Order SfM Optimization}
\label{sec:fastmap}

Following FastMap~\cite{fastmap}, we adopt a first-order SfM approach whose per-step cost is independent of 3D point count. Camera intrinsics are estimated via hierarchical interval search using a one-parameter division distortion model; focal length is recovered by maximizing the singular value ratio of the essential matrix. Global rotations are solved by minimizing geodesic distances on SO(3) using a continuous 6D parameterization, and translations are recovered via direction-only consistency with multiple random initializations.

\paragraph{Epipolar Adjustment}
The key step is structureless pose refinement using precomputed quadratic forms $W_n \in \mathbb{R}^{9 \times 9}$ from epipolar constraints:
\begin{equation}
    \mathcal{L}_e = \frac{2}{Z} \sum_{n=1}^{|\mathcal{P}|} \mathbf{e}_n^\top W_n \mathbf{e}_n,
\end{equation}
where $\mathbf{e}_n = \text{vec}(E_n)$ and $Z$ is a normalization constant. This enables BA-quality refinement with cost independent of point count, implemented with fused CUDA kernels.

\subsection{Importance-Guided MCMC Gaussian Allocation}
\label{sec:importance_guided}

\begin{figure}[t]
    \centering
    \includegraphics[width=\columnwidth]{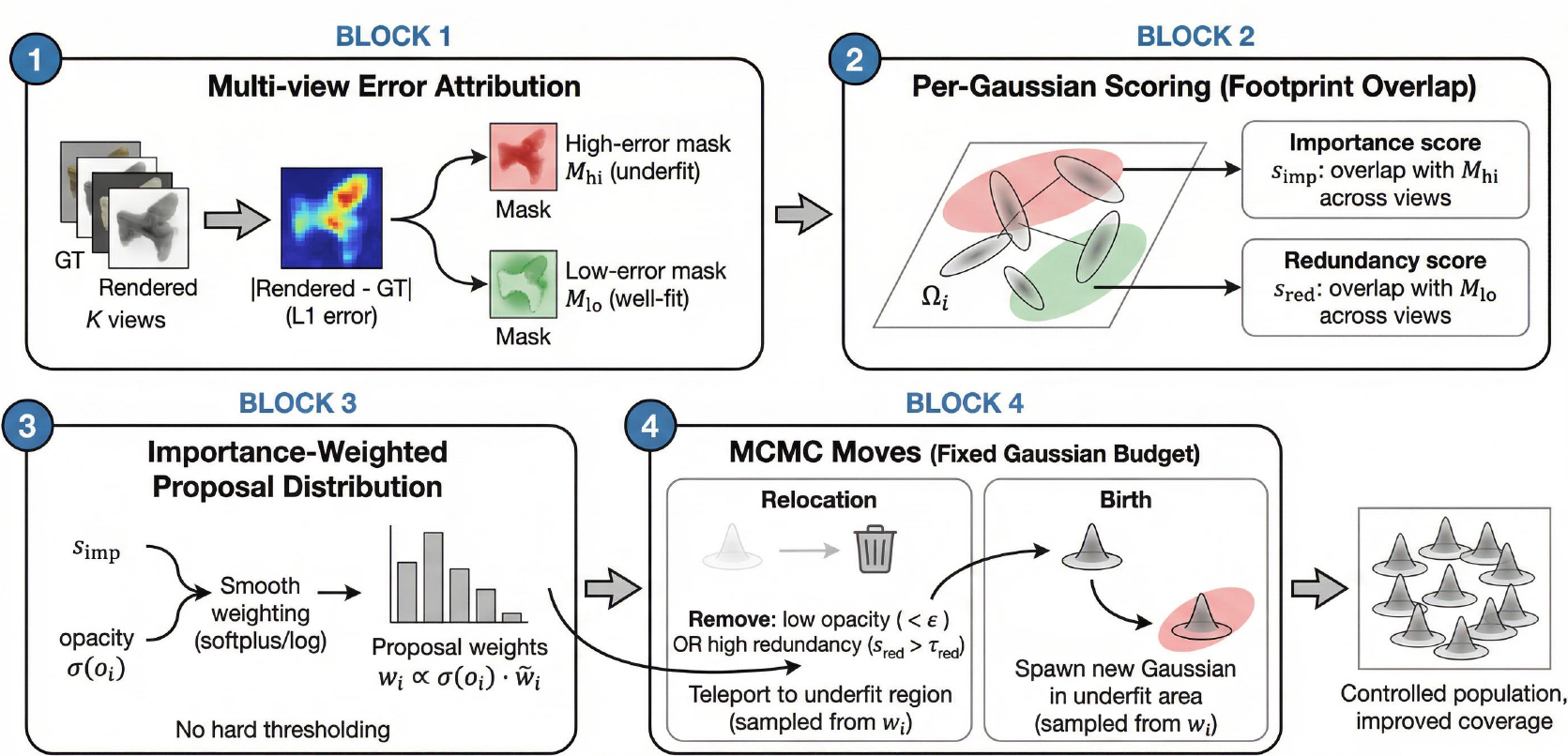}
    \caption{We aggregate multi-view reconstruction errors into per-Gaussian importance and redundancy scores, then use the resulting importance-weighted distribution for both \textbf{relocation} and \textbf{birth}. This reallocates model capacity toward persistently underfit regions while preserving the underlying MCMC population-management framework.}
    \label{fig:importance_mcmc}
\end{figure}

We adopt an MCMC-based approach to Gaussian population management (Figure~\ref{fig:importance_mcmc}), which differs fundamentally from standard 3DGS adaptive density control (ADC). While ADC uses gradient-based clone/split operations for densification and opacity thresholding for pruning, MCMC-based methods maintain a fixed or slowly-growing Gaussian budget through two operations: \textbf{relocation} (teleporting low-value Gaussians to new positions) and \textbf{birth} (adding new Gaussians by sampling from existing ones). We extend this framework with importance and redundancy scores derived from multi-view reconstruction error, and use them to define a smooth, importance-weighted sampling distribution that biases both operations toward underfit regions.

\paragraph{Multi-view Error Attribution}
Periodically during training, we sample $K$ views and compute per-pixel L1 error maps:
\begin{equation}
    e^j_{u,v} = \frac{1}{D} \sum_{d=1}^{D} |r^{j,d}_{u,v} - g^{j,d}_{u,v}|,
\end{equation}
where $r^j$ and $g^j$ are rendered and ground-truth images for view $j$, and $D{=}3$ is the RGB channel dimension. To make the signal robust across views and training stages, we avoid per-view min-max (which can make ``high-error'' non-selective late in training) and instead use \emph{robust quantile normalization}:
\begin{equation}
    \mathcal{M}^j = \mathrm{clip}\left(\frac{e^j - Q_{\ell}(e^j)}{Q_{h}(e^j)-Q_{\ell}(e^j)}, 0, 1\right),
\end{equation}
where $Q_{\ell}$ and $Q_{h}$ denote low/high quantiles (e.g., 5\% and 90\%). We then define per-view high/low thresholds by quantiles of $\mathcal{M}^j$:
\begin{equation}
    \tau^j_{\text{hi}} = Q_{q_{\text{hi}}}(\mathcal{M}^j), \quad \tau^j_{\text{lo}} = Q_{q_{\text{lo}}}(\mathcal{M}^j), \ \ (q_{\text{hi}} > q_{\text{lo}}).
\end{equation}

\paragraph{Importance Score (Underfit Attribution)}
For each Gaussian $\mathcal{G}_i$ with 2D footprint $\Omega_i^j$ in view $j$, we define a normalized underfit weight:
\begin{equation}
    \phi^j_{\text{hi}}(p)= \frac{\max(0, \mathcal{M}^j(p) - \tau^j_{\text{hi}})}{1 - \tau^j_{\text{hi}}}.
\end{equation}
We aggregate over the footprint and normalize by footprint size to reduce bias toward large projected Gaussians:
\begin{equation}
    s^i_{\text{imp}} = \frac{100}{K} \sum_{j=1}^{K} \frac{1}{|\Omega_i^j|} \sum_{p \in \Omega_i^j} \phi^j_{\text{hi}}(p).
\end{equation}
We report $s^i_{\text{imp}}$ on a 0--100 scale (percentage-like), so $\tau_{\text{imp}}$ is interpretable as the amount of persistent underfit mass required to prioritize a Gaussian. In practice, we estimate $|\Omega_i^j|$ using the projected radii from the renderer (cheap), rather than explicitly enumerating all pixels in $\Omega_i^j$.

\paragraph{Redundancy Score (Well-fit Attribution)}
We measure well-fit coverage analogously using a normalized low-error weight:
\begin{equation}
    \phi^j_{\text{lo}}(p)= \frac{\max(0, \tau^j_{\text{lo}} - \mathcal{M}^j(p))}{\tau^j_{\text{lo}}}.
\end{equation}
\begin{equation}
    s^i_{\text{red}} = \frac{1}{K} \sum_{j=1}^{K} \frac{1}{|\Omega_i^j|} \sum_{p \in \Omega_i^j} \phi^j_{\text{lo}}(p).
\end{equation}
We min-max normalize $s^i_{\text{red}}$ to $[0,1]$ for thresholding. Gaussians with high $s^i_{\text{red}}$ are redundant from a multi-view perspective---they occupy capacity in already well-reconstructed regions.

\paragraph{Importance-Weighted Sampling Distribution}
We convert importance into a smooth sampling weight:
\begin{equation}
    \begin{aligned}
    \tilde{w}_i &= \mathrm{softplus}\left(\frac{s^i_{\text{imp}}-\tau_{\text{imp}}}{\tau_{\text{imp}}}\right),\\
    w_i &\propto \left((1-\lambda_{\text{mix}})\sigma(o_i)+\lambda_{\text{mix}}\right)\tilde{w}_i.
    \end{aligned}
\end{equation}
where we normalize $\tilde{w}_i$ to have unit mean. The opacity mixing term $\lambda_{\text{mix}}$ prevents under-sampling Gaussians that are underfit but initially low-opacity (weak coverage), improving exploration beyond ``refine only already-visible'' regions. This replaces hard thresholding with a differentiable, scale-free weighting around $\tau_{\text{imp}}$.

\paragraph{Importance-Guided MCMC Strategy}
Our strategy extends the MCMC framework with an importance-weighted sampling distribution shared by both birth and relocation:
\begin{itemize}
    \item \textbf{Relocation}: Gaussians with low opacity ($< \epsilon$) or high redundancy score ($s^i_{\text{red}} > \tau_{\text{red}}$) are reassigned by resampling targets from $w_i$, recycling capacity from well-fit regions to underfit regions.
    \item \textbf{Birth}: New Gaussians are spawned by sampling parents from $w_i$, allocating new capacity to consistently underfit regions without hard thresholding.
\end{itemize}
Unlike standard ADC which can lead to unbounded Gaussian growth, this MCMC-based approach maintains controlled population while the importance guidance ensures capacity is directed toward multi-view consistent reconstruction.

This importance guidance replaces the uniform opacity-weighted sampling of 3DGS-MCMC with $w_i$; it is a heuristic allocation strategy---not a Metropolis--Hastings proposal---that leaves the underlying SGLD parameter updates unchanged and makes no additional convergence claims. Empirically, it substantially improves sample efficiency under fixed budgets (Table~\ref{tab:budget}).

\subsection{Joint Pose Optimization}
\label{sec:pose_opt}

While first-order SfM provides good initial poses, we jointly refine them during 3DGS training using both photometric and geometric losses.

\paragraph{Photometric Loss}
The primary supervision comes from image reconstruction:
\begin{equation}
    \mathcal{L}_{\text{photo}} = (1 - \lambda_s) \| \hat{I}(\mathbf{x}) - I(\mathbf{x}) \|_1 + \lambda_s (1 - \text{SSIM}(\hat{I}, I)),
\end{equation}
where $I$ and $\hat{I}$ are the ground-truth and rendered images, respectively, and SSIM denotes the Structural Similarity Index Measure. Camera poses $\{T_i\}$ are parameterized with learnable adjustments and optimized jointly with Gaussian parameters.

\paragraph{Bundle Adjustment Loss}
To anchor poses to geometric consistency, we incorporate a reprojection-based BA loss:
\begin{equation}
    \mathcal{L}_{\text{BA}} = \frac{1}{|\mathcal{O}|} \sum_{(i,j,k) \in \mathcal{O}} \| \pi(T_i, \mathbf{X}_k) - \mathbf{x}_{i,k} \|_2,
\end{equation}
where $\mathcal{O}$ denotes the set of 2D observations, $\mathbf{X}_k$ are triangulated 3D track points, $\pi$ is the projection function, and $\mathbf{x}_{i,k}$ is the observed 2D keypoint. The track points and their 2D associations are fixed at SfM initialization; only the camera poses and Gaussian parameters are optimized during joint training. This regularization prevents pose drift by maintaining consistency with sparse geometric constraints from SfM.

\paragraph{Combined Objective}
The total training loss combines all terms:
\begin{equation}
    \mathcal{L} = \mathcal{L}_{\text{photo}} + \lambda_{\text{BA}} \mathcal{L}_{\text{BA}} + \mathcal{L}_{\text{reg}},
\end{equation}
where $\mathcal{L}_{\text{reg}}$ includes standard 3DGS regularizers (scale, opacity). Camera 0 is fixed to resolve gauge ambiguity. This joint optimization allows the 3DGS representation and camera poses to co-evolve, leveraging photometric gradients for sub-pixel pose refinement while geometric constraints prevent degenerate solutions.

\section{Experimental Evaluation}
\label{sec:experiments}

\begin{table*}[t!]
    \centering
    \footnotesize
    \setlength\tabcolsep{1.5pt}
    \caption{
        \textbf{Quantitative comparisons.} Times include COLMAP preprocessing for prior methods. \colorbox{lightred}{\bf Best}, \colorbox{lightorange}{2nd}, \colorbox{lightyellow}{3rd}. Time in minutes.
    }
    \begin{tabular} {l | rrrrr | rrrrr | rrrrr}
        \toprule
        \multirow{2}{*}{Method}
        & \multicolumn{5}{c}{Mip-NeRF 360~\cite{barron2022mip}}  & \multicolumn{5}{c}{Deep Blending~\cite{hedman2018deep}} & \multicolumn{5}{c}{Tanks \& Temples~\cite{knapitsch2017tanks}} \\
        \cmidrule(l{2pt}r{2pt}){2-6} \cmidrule(l{2pt}r{2pt}){7-11} \cmidrule(l{2pt}r{2pt}){12-16}
         & Time$\downarrow$ & PSNR$\uparrow$ & SSIM$\uparrow$ & LPIPS$\downarrow$ & $\mathrm{N_{GS}}\downarrow$
        & Time$\downarrow$ & PSNR$\uparrow$ & SSIM$\uparrow$ & LPIPS$\downarrow$ & $\mathrm{N_{GS}}\downarrow$
        & Time$\downarrow$ & PSNR$\uparrow$ & SSIM$\uparrow$ & LPIPS$\downarrow$ & $\mathrm{N_{GS}}\downarrow$\\
        \midrule
        \multicolumn{16}{l}{\textit{w/o Pose Optimization}} \\
        \midrule
        3DGS~\cite{kerbl3Dgaussians}
            & 31.93 & 27.53 & 0.812 & 0.221 & 2.63M
            & 30.77 & 29.71 & 0.903 & \cellcolor{tabthird}0.241 & 2.46M
            & 22.34 & 23.71 & 0.850 & 0.170 & 1.57M\\
        3DGS-MCMC~\cite{mcmc3dgs}
            & 32.41 & \cellcolor{tabthird}28.01 & \cellcolor{tabthird}0.835 & 0.186 & 3.23M
            & 31.25 & \cellcolor{tabthird}29.78 & \cellcolor{tabfirst}\bf 0.912 & \cellcolor{tabsecond}0.237 & 2.95M
            & 22.89 & \cellcolor{tabsecond}24.40 & \cellcolor{tabfirst}\bf 0.869 & 0.149 & 1.85M\\
        Mini-Splatting~\cite{fang2024mini}
            & 28.69 & 27.32 & 0.821 & 0.217 & \cellcolor{tabsecond}0.53M
            & 24.35 & \cellcolor{tabsecond}29.99 & \cellcolor{tabsecond}0.907 & 0.244 & \cellcolor{tabthird}0.56M
            & 20.06 & 23.46 & 0.844 & 0.181 & \cellcolor{tabsecond}0.30M \\
        Speedy-splat~\cite{hanson2025speedy}
            & 24.38 & 26.91 & 0.781 & 0.295 & \cellcolor{tabfirst} \bf 0.30M
            & 21.75 & 29.42 & 0.898 & 0.272 & \cellcolor{tabfirst} \bf 0.25M
            & 17.32 & 23.38 & 0.816 & 0.242 & \cellcolor{tabfirst} \bf 0.18M \\
        Taming-3DGS~\cite{mallick2024taming}
            & \cellcolor{tabthird}16.36 & 27.48 & 0.794 & 0.261 & \cellcolor{tabthird}0.68M
            & \cellcolor{tabthird}14.06 & 29.50 & 0.894 & 0.278 & \cellcolor{tabsecond}0.29M
            & \cellcolor{tabthird}13.71 & 23.89 & 0.833 & 0.214 & \cellcolor{tabthird}0.32M \\
        DashGaussian~\cite{chen2025dashgaussian}
            & 17.35 & 27.73 & 0.817 & 0.218 & 2.40M
            & 15.16 & 29.65 & \cellcolor{tabthird}0.906 & 0.246 & 1.94M
            & 15.28 & 24.00 & \cellcolor{tabthird}0.853 & 0.178 & 1.21M \\
        FastGS-big~\cite{ren2025fastgs}
            & \cellcolor{tabsecond}14.58 & 27.93 & 0.820 & 0.216 & 1.15M
            & \cellcolor{tabsecond}13.00 & \cellcolor{tabfirst} \bf 30.12 & \cellcolor{tabsecond}0.907 & 0.243 & 0.65M
            & \cellcolor{tabsecond}13.03 & \cellcolor{tabthird}24.39 & \cellcolor{tabsecond}0.855 & 0.175 & 0.54M \\
        \midrule
        \multicolumn{16}{l}{\textit{w/ Pose Optimization}} \\
        \midrule
        GloSplat-A~\cite{glosplat}
            & 22.00 & \cellcolor{tabfirst}\bf 28.86 & \cellcolor{tabfirst} \bf 0.862 & \cellcolor{tabfirst}\bf 0.139 & 3.00M
            & 19.15 & 18.45$^*$ & 0.583$^*$ & 0.508$^*$ & 3.00M
            & 24.87 & 22.15 & 0.805 & \cellcolor{tabthird}0.147 & 3.00M \\
        VGGT-X~\cite{wang2025vggt}
            & 73.71 & 26.49 & 0.782 & \cellcolor{tabthird}0.177 & 3.00M
            & 57.24 & 18.25$^\dagger$ & 0.622$^\dagger$ & 0.545$^\dagger$ & 3.00M
            & 61.12 & 23.05 & 0.818 & \cellcolor{tabsecond}0.138 & 3.00M \\
        SalientGS (Ours)
            & \cellcolor{tabfirst}\bf 11.79 & \cellcolor{tabsecond}28.82 & \cellcolor{tabsecond}0.853 & \cellcolor{tabsecond}0.148 & 1.50M
            & \cellcolor{tabfirst}\bf 10.04 & 29.49 & \cellcolor{tabthird}0.906 & \cellcolor{tabfirst}\bf 0.183 & 1.50M
            & \cellcolor{tabfirst}\bf 10.03 & \cellcolor{tabfirst}\bf 24.65 & \cellcolor{tabfirst}\bf 0.869 & \cellcolor{tabfirst}\bf 0.109 & 1.50M \\
        \bottomrule
    \end{tabular}
    \\[2pt]
    {\scriptsize $^*$GloSplat-A and $^\dagger$VGGT-X consistently fail on the \textit{drjohnson} scene, significantly degrading their averaged Deep Blending metrics.}
    \label{tab:main-result}
\end{table*}

\subsection{Implementation Details}
\label{sec:impl_details}

We evaluate on three standard benchmarks: Mip-NeRF 360~\cite{barron2022mip} (9 scenes), Deep Blending~\cite{hedman2018deep} (2 scenes), and Tanks \& Temples~\cite{knapitsch2017tanks} (2 scenes). We report scene-averaged PSNR, SSIM, and LPIPS~\cite{zhang2018unreasonable}, end-to-end time (feature extraction, matching, SfM, and training), and Gaussian count $\mathrm{N_{GS}}$. Scene averaging matches the aggregation used for the comparison methods; pooled per-image metrics are reported separately in the supplementary material and are not mixed into Table~\ref{tab:main-result}. To summarize cross-dataset behavior, we additionally macro-average the three dataset-level entries, giving each benchmark equal weight. Failed scenes remain in their dataset aggregate; in particular, the \textit{drjohnson} failures of GloSplat-A and VGGT-X are not dropped. The released-code verification uses seed 42 and a single NVIDIA RTX PRO 6000 Blackwell GPU. For Fisher Vector encoding, we train a GMM with $M{=}64$ components on SIFT descriptors and retrieve the top-$k{=}20$ candidates per image. The Gaussian budget cap is set to $\mathrm{N_{GS}}{=}1.5$M. For importance-guided MCMC, we use robust normalization quantiles $(\ell,h){=}(0.05,0.90)$, high/low selection quantiles $(q_{\text{hi}},q_{\text{lo}}){=}(0.9,0.1)$, importance threshold $\tau_{\text{imp}}{=}5$ (on a 0--100 importance scale), redundancy threshold $\tau_{\text{red}}{=}0.9$, opacity mixing $\lambda_{\text{mix}}{=}0.05$, and $K{=}10$ views for score computation. Joint training runs for exactly 30K iterations with $\lambda_{\text{BA}}{=}0.01$ and $\lambda_s{=}0.2$. Importance scores are first computed after a pose warmup of 3K iterations and recomputed every $T{=}500$ iterations thereafter.

\subsection{Main Results}
\label{sec:main_results}

Table~\ref{tab:main-result} compares SalientGS against seven methods that rely on COLMAP preprocessing (upper block) and two concurrent pose-optimizing methods, GloSplat-A~\cite{glosplat} and VGGT-X~\cite{wang2025vggt} (lower block). All reported times include COLMAP preprocessing for prior methods to reflect true end-to-end cost.

\paragraph{Quality}
SalientGS leads the three-benchmark macro-average in PSNR (27.65~dB), SSIM (0.876), and LPIPS (0.147). It is also the only method in Table~\ref{tab:main-result} with top-two LPIPS on every benchmark. On Mip-NeRF 360 it obtains 28.82~dB, 0.853 SSIM, and 0.148 LPIPS with 1.5M Gaussians; its LPIPS is 20.4\% below 3DGS-MCMC and 31.5\% below FastGS-big. On Deep Blending, it records 29.49~dB / 0.906 / 0.183 and reconstructs \textit{drjohnson}, where GloSplat-A and VGGT-X fail. On Tanks \& Temples, it leads PSNR (24.65) and LPIPS (0.109) and ties the best SSIM (0.869). The macro-average thus measures consistency rather than one favorable dataset.

\paragraph{Efficiency and Model Size}
Including every front-end and training stage, SalientGS averages 11.79, 10.04, and 10.03 minutes on the three benchmarks, respectively. It is the fastest end-to-end method in each block of Table~\ref{tab:main-result}; on Mip-NeRF 360 it is 1.87$\times$ faster than GloSplat-A and 6.25$\times$ faster than VGGT-X, with half their Gaussian count. The released CSV and runner make this hardware-dependent comparison reproducible.

\subsection{Ablation Study}
\label{sec:ablation}

We ablate each key component of SalientGS on the Mip-NeRF 360 benchmark to quantify individual contributions. All variants share the same first-order SfM initialization and Fisher Vector matching; only the 3DGS training stage differs unless otherwise noted. A previous evaluation error affected only the reported full-model aggregate; after correction, the full model is 28.82~dB / 0.853 / 0.148 under the same protocol as the verified variants. We use this corrected value consistently, and all deltas below are recomputed against it.

\begin{table}[t]
    \centering
    \scriptsize
    \setlength\tabcolsep{1.5pt}
    \renewcommand{\arraystretch}{0.95}
    \caption{
        \textbf{Component ablation on Mip-NeRF 360.} $\Delta$PSNR is relative to the full configuration.
    }
    \begin{tabular}{l ccc r}
        \toprule
        Configuration & PSNR$\uparrow$ & SSIM$\uparrow$ & LPIPS$\downarrow$ & $\Delta$PSNR \\
        \midrule
        \multicolumn{5}{l}{\textit{(A) Importance-Guided MCMC}} \\
        \quad Full model & \textbf{28.82} & \textbf{0.853} & \textbf{0.148} & --- \\
        \quad w/o Guided Birth & 28.72 & 0.847 & 0.153 & $-$0.10 \\
        \quad w/o Guided Reloc. & 28.77 & 0.849 & 0.149 & $-$0.05 \\
        \quad w/o Both (Vanilla MCMC) & 28.72 & 0.848 & 0.149 & $-$0.10 \\
        \quad w/o Footprint Normalization & 22.70 & 0.684 & 0.422 & $-$6.12 \\
        \midrule
        \multicolumn{5}{l}{\textit{(B) Densification Strategy}} \\
        \quad Standard ADC (clone/split) & 27.67 & 0.828 & 0.162 & $-$1.15 \\
        \bottomrule
    \end{tabular}
    \label{tab:ablation}
\end{table}

\paragraph{Importance-Guided MCMC (Group A)}
Against vanilla MCMC, guidance adds 0.10~dB and lowers LPIPS from 0.149 to 0.148. Birth guidance provides the larger PSNR contribution; relocation is most useful when birth is also guided. Removing footprint normalization costs 6.12~dB because large projected primitives otherwise dominate the scores.

\paragraph{Densification Strategy (Group B)}
Replacing guided MCMC management with standard ADC costs 1.15~dB, supporting multi-view redistribution under a fixed budget.

\paragraph{Joint Pose Optimization and SfM Initialization}
Table~\ref{tab:ablation_sfm} presents ablations on joint optimization and SfM initialization quality.

\begin{table}[t]
    \centering
    \scriptsize
    \setlength\tabcolsep{1.5pt}
    \caption{
        \textbf{Pose/SfM ablation} on Mip-NeRF 360. $\Delta$PSNR is relative to the full configuration.
    }
    \begin{tabular}{l ccc r}
        \toprule
        Configuration & PSNR$\uparrow$ & SSIM$\uparrow$ & LPIPS$\downarrow$ & $\Delta$PSNR \\
        \midrule
        \multicolumn{5}{l}{\textit{(A) Joint Pose Optimization}} \\
        \quad Full model & \textbf{28.82} & \textbf{0.853} & \textbf{0.148} & --- \\
        \quad w/o BA Loss (photometric-only) & 28.70 & 0.845 & 0.156 & $-$0.12 \\
        \quad Freeze Poses after SfM Init & 28.32 & 0.836 & 0.160 & $-$0.50 \\
        \midrule
        \multicolumn{5}{l}{\textit{(B) Sensitivity to SfM Initialization Quality}} \\
        \quad Higher $k{=}40$ retrieval & \textbf{28.86} & \textbf{0.860} & \textbf{0.134} & $+$0.04 \\
        \quad Full model ($k{=}20$ retrieval) & 28.82 & 0.853 & 0.148 & --- \\
        \quad Lower $k{=}10$ retrieval & 27.35 & 0.833 & 0.169 & $-$1.47 \\
        \quad Lower $k{=}5$ retrieval$^\dagger$ & 26.57 & 0.774 & 0.257 & $-$2.25 \\
        \bottomrule
    \end{tabular}
    \\[2pt]
    {\scriptsize $^\dagger$At $k{=}5$, 1/9 scenes fails catastrophically (stump: 14.97~dB PSNR), significantly degrading the average.}
    \label{tab:ablation_sfm}
\end{table}

\paragraph{Joint Pose Optimization (Group A)}
Removing BA costs 0.12~dB, whereas freezing poses costs 0.50~dB. Thus photometric refinement provides most of the recovery and reprojection anchoring adds a further measurable gain.

\paragraph{SfM Initialization Quality (Group B)}
Increasing retrieval from $k{=}20$ to 40 gives a small 0.04~dB gain, whereas reducing it to 10 and 5 costs 1.47 and 2.25~dB, respectively; at $k{=}5$, \textit{stump} fails at 14.97~dB. Joint optimization therefore cannot replace a connected, reliable view graph.

\begin{table}[t]
    \centering
    \footnotesize
    \setlength\tabcolsep{4pt}
    \caption{
        \textbf{Gaussian budget efficiency on Mip-NeRF 360.} PSNR at varying $\mathrm{N_{GS}}$ caps for vanilla MCMC vs.\ importance-guided MCMC. $\Delta$ shows the gain from importance guidance.
    }
    \begin{tabular}{r cc c}
        \toprule
        $\mathrm{N_{GS}}$ Cap & Vanilla MCMC & Guided MCMC & $\Delta$PSNR \\
        \midrule
        500K  & 28.19 & 28.46 & +0.27 \\
        1.0M  & 28.59 & 28.81 & +0.22 \\
        1.5M  & 28.72 & 28.82 & +0.10 \\
        2.0M  & 28.80 & 28.93 & +0.13 \\
        3.0M  & 28.88 & \textbf{28.97} & +0.09 \\
        \bottomrule
    \end{tabular}
    \label{tab:budget}
\end{table}

\paragraph{Gaussian Budget Efficiency}
Guidance improves every tested budget, with the gain generally decreasing from +0.27~dB at 500K to +0.09~dB at 3M. Guided 1M (28.81~dB) already exceeds vanilla 1.5M (28.72~dB), showing the largest value under tight capacity.

\paragraph{Schedule and normalization sensitivity}
The supplementary material reports every schedule and normalization setting. The released defaults were fixed before the full benchmark rather than selected per scene for the best metric: $T{=}500$ reduces score-recomputation frequency relative to $T{=}200$, $k{=}20$ limits retrieval and matching cost relative to $k{=}40$, and one quantile setting is retained across all scenes. Some more expensive or sweep-selected alternatives achieve better individual metrics, which we report without substituting them into the released result.

\subsection{SfM Runtime Comparison}
\label{sec:sfm_runtime}

\begin{figure}[t]
    \centering
    \includegraphics[width=\columnwidth]{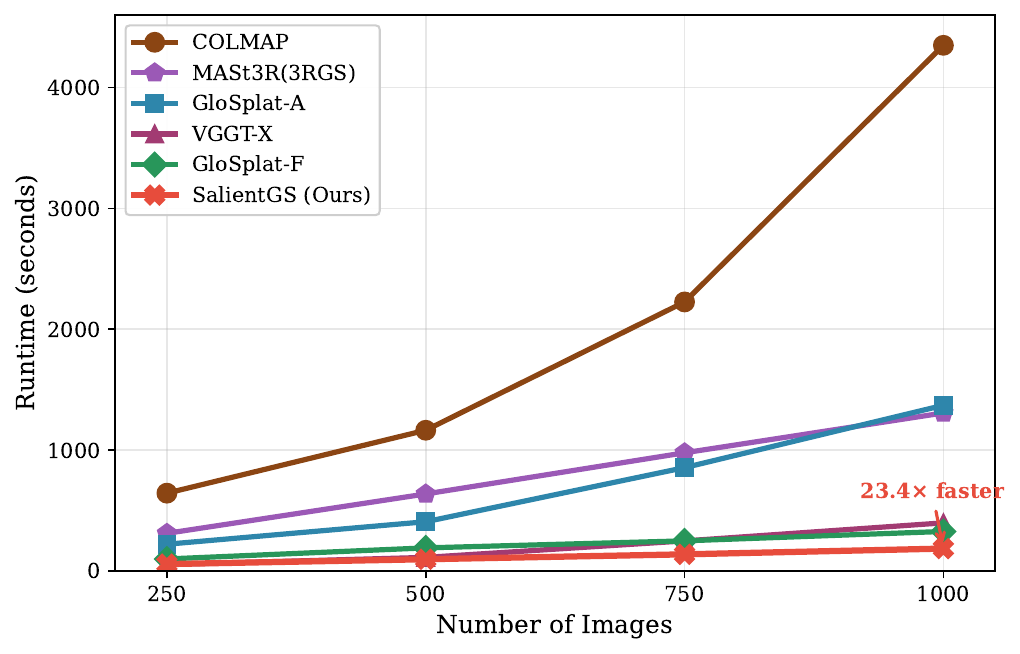}
    \caption{SfM runtime vs.\ image count. SalientGS achieves up to 23$\times$ speedup over COLMAP with near-linear scaling.}
    \label{fig:runtime_comparison}
\end{figure}

Figure~\ref{fig:runtime_comparison} compares SfM runtime on the Courthouse scene (250--1000 images) against COLMAP, MASt3R~\cite{leroy2024grounding}, VGGT~\cite{wang2025vggt}, and GloSplat~\cite{glosplat}. SalientGS achieves 13.5--23.4$\times$ speedup over COLMAP with near-linear scaling; the full numeric breakdown is provided in the Supplementary ``SfM Runtime Details'' section.

\subsection{Qualitative Results}
\label{sec:qualitative}

\begin{figure*}[!htbp]
    \centering
    \includegraphics[width=0.49\textwidth]{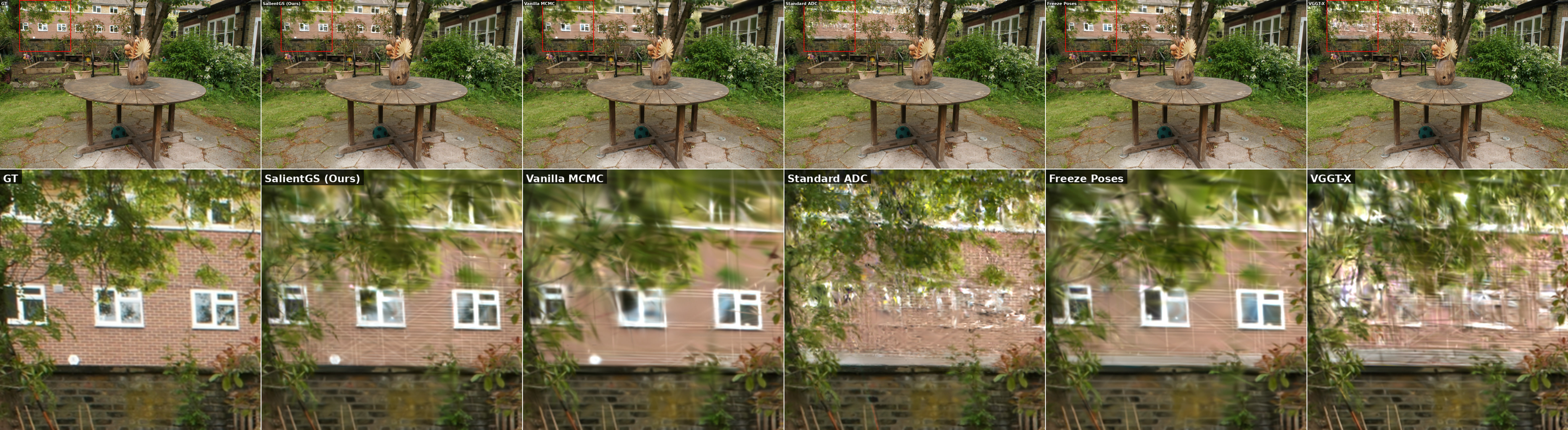}\hfill
    \includegraphics[width=0.49\textwidth]{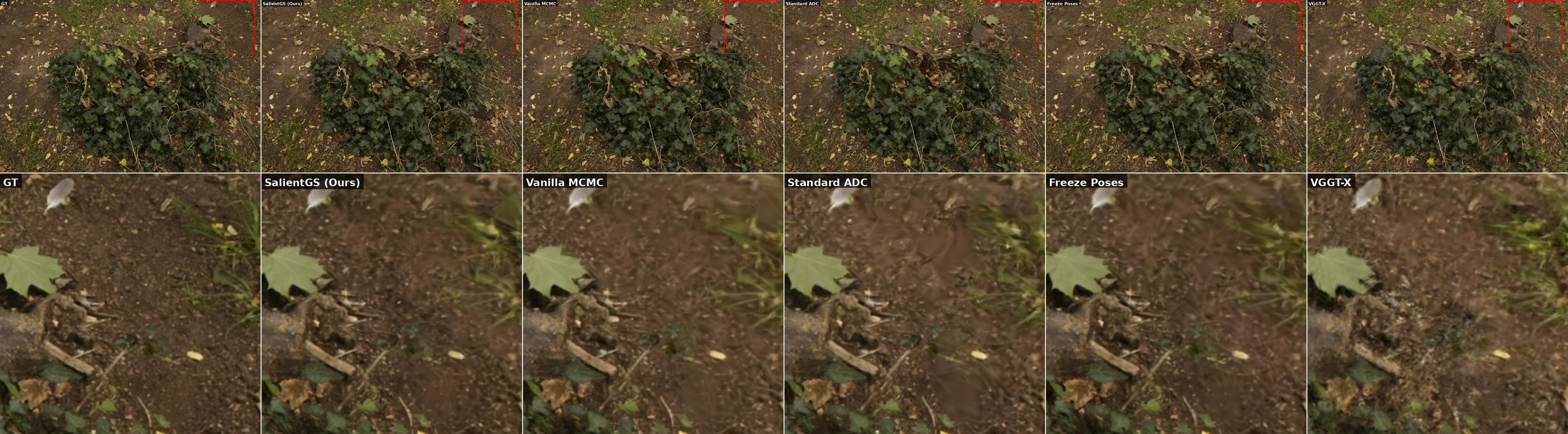}\\[2pt]
    \includegraphics[width=0.49\textwidth]{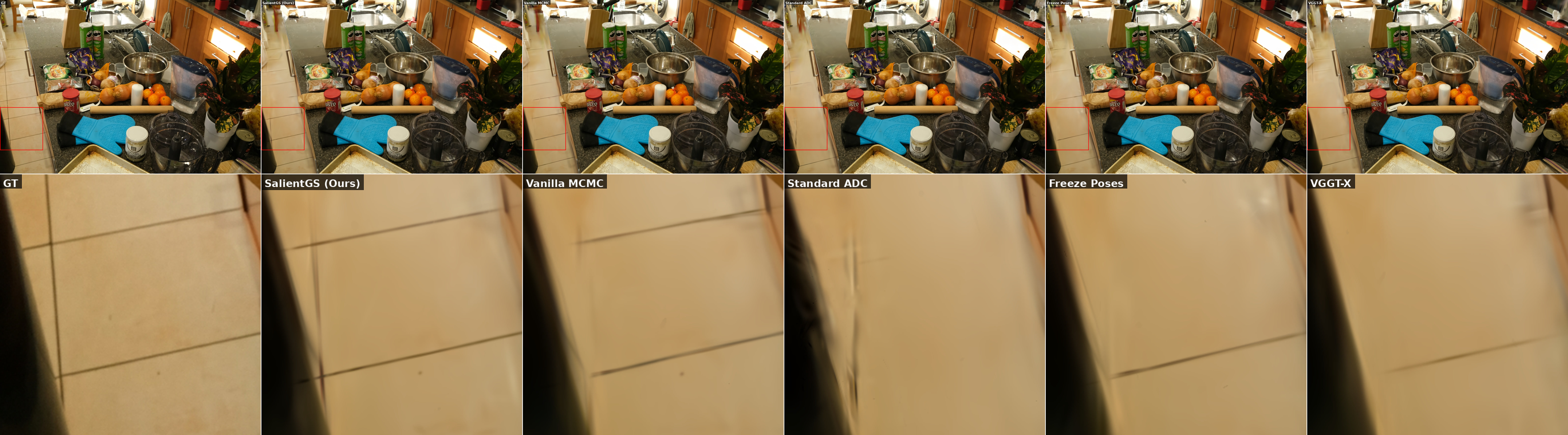}\hfill
    \includegraphics[width=0.49\textwidth]{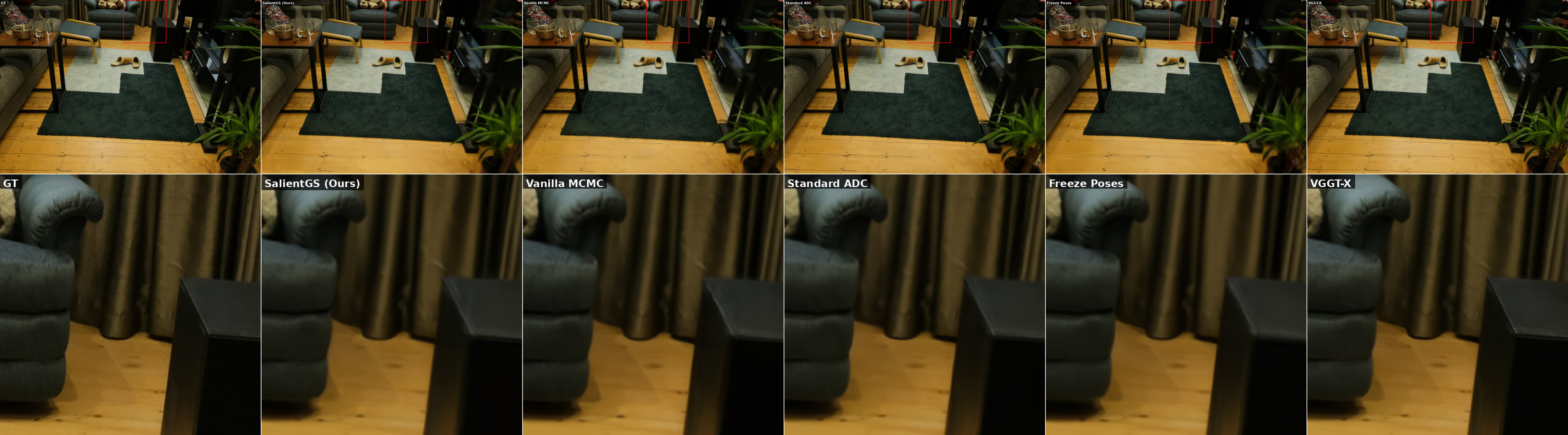}
    \caption{\textbf{Selected-view qualitative comparison on Mip-NeRF 360.} The first three panels illustrate localized detail differences on \textit{garden}, \textit{stump}, and \textit{counter}; they are not used to claim per-scene ranking. The final panel is a selected \textit{room} failure view, where SalientGS has per-view LPIPS 0.312 versus 0.244 for VGGT-X. Scene-mean LPIPS is reported separately in the supplementary material.}
    \label{fig:qual_main}
\end{figure*}

We present qualitative comparisons on representative Mip-NeRF 360 scenes in Figure~\ref{fig:qual_main}. For each view, we show the full rendered image alongside a zoomed crop of the region with the greatest perceptual difference (highest LPIPS gap) between our method and the baselines.

\paragraph{Selected-View Observations}
The crops illustrate where importance-guided allocation can recover localized high-frequency structure. The most consistent quantitative gain among these examples is on \textit{stump}. On the selected \textit{garden} and \textit{counter} views, SalientGS preserves visually sharp local detail, but its scene-mean LPIPS does not lead every ablation: Standard ADC is better on \textit{garden}, while Vanilla MCMC and Freeze Poses are slightly better on \textit{counter}. We therefore use these panels as qualitative diagnostics rather than as evidence of scene-level ranking.

\paragraph{Failure Cases}
On the selected texture-poor \textit{room} view, VGGT-X's dense feed-forward initialization provides a stronger local reconstruction than our first-order SfM (per-view LPIPS 0.244 vs.\ 0.312). This is a localized failure rather than the scene-average ranking: SalientGS records 0.150 scene-mean LPIPS on \textit{room}. Detailed per-scene and per-view distinctions are provided in the Supplementary ``Per-Image LPIPS Analysis'' and failure-diagnostics sections.

\subsection{Pose Evaluation on ETH3D SLAM}
\label{sec:pose_eval}

To evaluate pose accuracy independently of rendering quality, we benchmark on 45 training sequences from the ETH3D SLAM dataset~\cite{Schops_2019_CVPR}, which provides millimeter-accurate ground-truth poses from motion capture. These sequences span 17 scene groups covering sparse textures, dynamic objects, and drastic illumination changes.
Following the GLOMAP~\cite{pan2024glomap} evaluation protocol, we report absolute camera position accuracy after robust Procrustes alignment: \textbf{Recall@0.1m} (fraction of cameras within 10\,cm of ground truth), and \textbf{AUC} at 0.1\,m and 0.5\,m thresholds.

This experiment is designed to isolate the contribution of joint optimization of pose and appearance from SfM initialization quality. We evaluate three SfM methods---\textbf{FastMap}~\cite{fastmap}, \textbf{COLMAP}~\cite{colmap}, and \textbf{GLOMAP}~\cite{pan2024glomap}---each in two configurations: (1)~poses used directly without refinement (init), and (2)~poses after joint training with photometric and BA losses (\S\ref{sec:pose_opt}). This cross-product design reveals both the effect of initialization quality and the consistent benefit of joint refinement.

\begin{table}[t]
    \centering
    \footnotesize
    \setlength\tabcolsep{3pt}
    \caption{
        \textbf{Pose evaluation on ETH3D SLAM~\cite{Schops_2019_CVPR}.}
        Three SfM initializations---FastMap~\cite{fastmap}, COLMAP~\cite{colmap}, GLOMAP~\cite{pan2024glomap}---evaluated before and after joint optimization (\S\ref{sec:pose_opt}).
        Results on two subsets: 39 sequences where SfM and joint training both succeed, and 34 where 3DGS reconstruction also succeeds (PSNR${>}$20).
    }
    \begin{tabular}{@{}llccc@{}}
        \toprule
        Subset & Method & \shortstack[c]{Recall@\\0.1m} & \shortstack[c]{AUC@\\0.1m} & \shortstack[c]{AUC@\\0.5m} \\
        \midrule
        \multirow{6}{*}{\shortstack[l]{Train-succ.\\(39 seq.)}}
        & FastMap (init) & 50.5 & 37.2 & 58.2 \\
        & FastMap + joint opt & \textbf{57.0} & \textbf{42.8} & \textbf{61.0} \\
        & COLMAP (init) & 50.2 & 38.5 & 59.8 \\
        & COLMAP + joint opt & 52.0 & 40.5 & 60.4 \\
        & GLOMAP (init) & 55.5 & 40.1 & 60.2 \\
        & GLOMAP + joint opt & 56.5 & 42.3 & 60.7 \\
        \midrule
        \multirow{6}{*}{\shortstack[l]{Recon.\\succ.\\(34 seq.)}}
        & FastMap (init) & 55.8 & 41.0 & 61.2 \\
        & FastMap + joint opt & \textbf{62.0} & \textbf{47.5} & \textbf{64.5} \\
        & COLMAP (init) & 61.0 & 40.8 & 60.3 \\
        & COLMAP + joint opt & 61.2 & 42.5 & 61.0 \\
        & GLOMAP (init) & 60.2 & 44.5 & 63.5 \\
        & GLOMAP + joint opt & 61.4 & 47.0 & 64.1 \\
        \bottomrule
    \end{tabular}
    \label{tab:eth3d_slam}
\end{table}

Table~\ref{tab:eth3d_slam} presents the results. Of our 45 sequences, SfM successfully registers cameras in 43 (96\%); 39 of these also complete joint training (Train-succ.), and 34 achieve full reconstruction success (Recon.\ succ., PSNR${>}$20).

Joint optimization consistently improves all pose metrics regardless of SfM initialization. The improvement is largest for FastMap, the weakest initialization: AUC@0.1m improves by +5.6 (39~seq.) and +6.5 (34~seq.), compared to +2.0/+1.7 for COLMAP and +2.2/+2.5 for GLOMAP. Strikingly, FastMap + joint optimization achieves the best pose accuracy across all metrics on both subsets, surpassing even GLOMAP + joint optimization (e.g., AUC@0.1m 42.8 vs.\ 42.3 on 39~seq.; 47.5 vs.\ 47.0 on 34~seq.). We attribute this to the nature of first-order SfM errors: they are smooth and systematic rather than discrete outliers, making them particularly amenable to correction via photometric and reprojection gradients. This result validates the SalientGS pipeline design---fast first-order SfM initialization paired with joint refinement not only matches but slightly exceeds the pose accuracy of stronger SfM baselines.

\section{Conclusion and Limitations}
\label{sec:conclusion}

We presented SalientGS, a unified SfM-to-3DGS pipeline for fast reconstruction with robust quality under a fixed Gaussian budget. In the released-code 13-scene verification, SalientGS achieves the best three-benchmark macro-average PSNR, SSIM, LPIPS, and end-to-end runtime among all methods in the main comparison. This overall result is driven by consistency: SalientGS remains strong across all three datasets, whereas competing pose-optimizing pipelines suffer severe failures on Deep Blending. The result supports the practical value of combining retrieval-based matching, first-order SfM, joint pose refinement, and importance-guided allocation. At the same time, individual dataset metrics show room for improvement, particularly Deep Blending PSNR/SSIM and fine-detail LPIPS on several Mip-NeRF 360 scenes.

\paragraph{Design Choice: Minimal Per-Scene Tuning}
SalientGS uses a single fixed set of hyperparameters across all scenes, unlike methods that use per-scene schedules~\cite{mcmc3dgs,ren2025fastgs}. The defaults are an efficiency and reproducibility choice, not the metric-wise optimum of the post-hoc sweep: more frequent score updates, broader retrieval, or alternative quantiles can improve individual metrics. Retaining one preselected configuration simplifies deployment and makes the released runner deterministic under a fixed seed, but it leaves some quality on the table (see the Supplementary sensitivity tables).

\paragraph{Limitation: Pose Optimization vs.\ Speed}
A fundamental tension exists between camera pose optimization and training efficiency. The verified end-to-end runtime is competitive, but joint optimization still adds training-stage overhead relative to a frozen-pose run on identical hardware. Reducing that overhead without weakening geometric correction remains open.

\paragraph{Limitation: Upstream Component Dependencies}
Joint optimization is effective only when upstream SfM succeeds: of 45 ETH3D SLAM sequences, 2 fail SfM entirely and 4 fail joint training. Across the 39 training-success sequences, aggregate pose metrics improve, while per-scene AUC@0.1m improves in 67\% of cases. Additionally, Fisher Vector retrieval may fail under severe appearance variation where learned descriptors would provide more robust pair selection, and first-order SfM is sensitive to sparse coverage and degenerate camera motions such as collinear configurations. These limitations highlight that our joint optimization refines rather than substitutes for successful SfM initialization.

\bibliographystyle{ACM-Reference-Format}
\bibliography{references}

\end{document}